\documentclass[sigconf]{acmart}
\renewcommand\footnotetextcopyrightpermission[1]{} % removes footnote with conference information in first column

\usepackage{booktabs} % For formal tables
\usepackage{url}
\usepackage{amsmath}
\usepackage{amssymb}
\usepackage{dirtytalk}
\usepackage{graphicx}
\usepackage{hyperref}
\usepackage{wrapfig}
\renewcommand{\vec}[1]{\mathbf{#1}}

\usepackage{listings}
\usepackage{color}

\definecolor{dkgreen}{rgb}{0,0.6,0}
\definecolor{gray}{rgb}{0.5,0.5,0.5}
\definecolor{mauve}{rgb}{0.58,0,0.82}

\setcopyright{none}

\acmConference[]{}{}{}
\acmYear{2017}
\copyrightyear{2017}

\begin{document}
\title{An Architecture Combining Convolutional Neural Network (CNN) and Support Vector Machine (SVM) for Image Classification}

\author{Abien Fred M. Agarap}
\email{abienfred.agarap@gmail.com}

\begin{abstract}
Convolutional neural networks (CNNs) are similar to ``ordinary'' neural networks in the sense that they are made up of
hidden layers consisting of neurons with ``learnable'' parameters. These neurons receive inputs, performs a dot product, and
then follows it with a non-linearity. The whole network expresses the mapping between raw image pixels and their class scores.
Conventionally, the Softmax function is the classifier used at the last layer of this network. However, there have been studies \cite{alalshekmubarak2013novel, agarap2017neural, Tang} conducted to challenge this norm.
The cited studies introduce the usage of linear support vector machine (SVM) in an artificial neural network architecture. This project is yet another take on the subject, and is inspired by \cite{Tang}. Empirical data has shown that the CNN-SVM model was able to achieve a test accuracy of $\approx$99.04\% using the MNIST dataset\cite{lecun2010mnist}. On the other hand, the CNN-Softmax was able to achieve a test accuracy of $\approx$99.23\% using the same dataset. Both models were also tested on the recently-published Fashion-MNIST dataset\cite{xiao2017fashion}, which is suppose to be a more difficult image classification dataset than MNIST\cite{zalandoresearch_2017}. This proved to be the case as CNN-SVM reached a test accuracy of $\approx$90.72\%, while the CNN-Softmax reached a test accuracy of $\approx$91.86\%. The said results may be improved if data preprocessing techniques were employed on the datasets, and if the base CNN model was a relatively more sophisticated than the one used in this study.
\end{abstract}




\maketitle
\section{Introduction}
A number of studies involving deep learning approaches have claimed state-of-the-art performances in a considerable number of tasks. These include, but are not limited to, image classification\cite{krizhevsky2012imagenet}, natural language processing\cite{wen2015semantically}, speech recognition\cite{chorowski2015attention}, and text classification\cite{yang2016hierarchical}. The models used in the said tasks employ the softmax function at the classification layer.

However, there have been studies\cite{agarap2017neural, alalshekmubarak2013novel, Tang} conducted that takes a look at an alternative to softmax function for classification -- the support vector machine (SVM). The aforementioned studies have claimed that the use of SVM in an artificial neural network (ANN) architecture produces a relatively better results than the use of the conventional softmax function. Of course, there is a drawback to this approach, and that is the restriction to binary classification. As SVM aims to determine the optimal hyperplane separating two classes in a dataset, a multinomial case is seemingly ignored. With the use of SVM in a multinomial classification, the case becomes a one-versus-all, in which the positive class represents the class with the highest score, while the rest represent the negative class.

In this paper, we emulate the architecture proposed by \cite{Tang}, which combines a convolutional neural network (CNN) and a linear SVM for image classification. However, the CNN employed in this study is a simple 2-Convolutional Layer with Max Pooling model, in contrast with the relatively more sophisticated model and preprocessing in \cite{Tang}.

\section{Methodology}

\subsection{Machine Intelligence Library}
Google TensorFlow\cite{tensorflow2015-whitepaper} was used to implement the deep learning algorithms in this study.

\subsection{The Dataset}
MNIST\cite{lecun2010mnist} is an established \textit{standard} handwritten digit classification dataset that is widely used for benchmarking deep learning models. It is a 10-class classification problem having 60,000 training examples, and 10,000 test cases -- all in grayscale. However, it is argued that the MNIST dataset is ``too easy'' and ``overused'', and ``it can not represent modern CV [Computer Vision] tasks''\cite{zalandoresearch_2017}. Hence, \cite{xiao2017fashion} proposed the Fashion-MNIST dataset. The said dataset consists of Zalando's article images having the same distribution, the same number of classes, and the same color profile as MNIST.
\begin{table}[!htb]
\centering
\caption{Dataset distribution for both MNIST and Fashion-MNIST.}
		\begin{tabular}{ccc}
		\toprule
		Dataset & MNIST & Fashion-MNIST \\
		\midrule
		Training & 60,000 & 10,000 \\
		Testing & 60,000 & 10,000 \\
		\bottomrule
		\end{tabular}\\
		\label{table: distribution}
\end{table}

\indent	Both datasets were used as they were, with no preprocessing such as normalization or dimensionality reduction.

\subsection{Support Vector Machine (SVM)}
The support vector machine (SVM) was developed by Vapnik\cite{Cortes} for binary classification. Its objective is to find the optimal hyperplane $f(\vec{w}, \vec{x}) = \vec{w} \cdot \vec{x} + b$ to separate two classes in a given dataset, with features $\vec{x} \in \mathbb{R}^{m}$.

SVM learns the parameters $\vec{w}$ by solving an optimization problem (Eq. \ref{l1-svm}).
\begin{equation} \label{l1-svm}
min \dfrac{1}{p}\vec{w}^{T}\vec{w} + C \sum_{i = 1}^{p} max\big(0, 1 - y_{i}'(\vec{w}^{T}\vec{x}_{i}+b)\big)
\end{equation}

where $\vec{w}^{T} \vec{w}$ is the Manhattan norm (also known as L1 norm), $C$ is the penalty parameter (may be an arbitrary value or a selected value using hyper-parameter tuning), $\vec{y'}$ is the actual label, and $\vec{w}^T\vec{x} + b$ is the predictor function. Eq. \ref{l1-svm} is known as L1-SVM, with the standard hinge loss. Its differentiable counterpart, L2-SVM (Eq. \ref{l2-svm}), provides more stable results\cite{Tang}.
\begin{equation}\label{l2-svm}
min \dfrac{1}{p}\|\vec{w}\|_{2}^{2} + C \sum_{i = 1}^{p} max\big(0, 1 - y_{i}'(\vec{w}^{T}\vec{x}_{i}+b)\big)^{2}
\end{equation}

where $\|\vec{w}\|_{2}$ is the Euclidean norm (also known as L2 norm), with the squared hinge loss.

\subsection{Convolutional Neural Network (CNN)}
Convolutional Neural Network (CNN) is a class of deep feed-forward artificial neural networks which is commonly used in computer vision problems such as image classification. The distinction of CNN from a ``plain'' multilayer perceptron (MLP) network is its usage of convolutional layers, pooling, and non-linearities such as $tanh$, $sigmoid$, and ReLU.

The convolutional layer (denoted by \texttt{CONV}) consists of a filter, for instance, $5 \times 5 \times 1$ (5 pixels for width and height, and 1 because the images are in grayscale). Intuitively speaking, the \texttt{CONV} layer is used to ``slide'' through the width and height of an input image, and compute the dot product of the input's region and the \texttt{weight} learning parameters. This in turn will produce a 2-dimensional activation map that consists of responses of the filter at given regions.

Consequently, the pooling layer (denoted by \texttt{POOL}) reduces the size of input images as per the results of a \texttt{CONV} filter. As a result, the number of parameters within the model is also reduced -- called \textit{down-sampling}.

\begin{figure}[htb!]
\minipage{0.45\textwidth}
\centering
	\includegraphics[width=\linewidth]{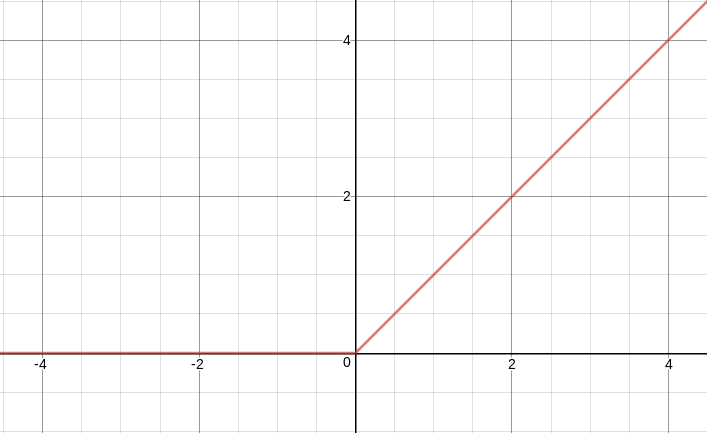}
	\caption{The Rectified Linear Unit (ReLU) activation function produces $0$ as an output when $x < 0$, and then produces a linear with slope of $1$ when $x > 0$.}
	\label{relu-graph}
\endminipage\hfill
\end{figure}

Lastly, an activation function is used for introducing non-linearities in the computation. Without such, the model will only learn linear mappings. The commonly-used activation function these days is the ReLU function\cite{hahnloser2000digital} (see Figure \ref{relu-graph}). ReLU is commonly-used over $tanh$ and $sigmoid$ for it was found out that it greatly accelerates the convergence of stochastic gradient descent compared the other two functions\cite{krizhevsky2012imagenet}. Furthermore, compared to the extensive computation required by $tanh$ and $sigmoid$, ReLU is implemented by simply thresholding matrix values at zero (see Eq. \ref{relu}).
\begin{equation}\label{relu}
f\big(h_{\theta}(x)\big) = h_{\theta}(x)^{+} = max\big(0, h_{\theta}(x)\big)
\end{equation}

In this paper, we implement a base CNN model with the following architecture:
\begin{enumerate}
\item INPUT: $32 \times 32 \times 1$
\item {\color{green}CONV5: $5 \times 5$ size, 32 filters, 1 stride}
\item {\color{red}ReLU: $max(0, h_{\theta}(x))$}
\item {\color{blue}POOL: $2 \times 2$ size, 1 stride}
\item {\color{green}CONV5: $5 \times 5$ size, 64 filters, 1 stride}
\item {\color{red}ReLU: $max(0, h_{\theta}(x))$}
\item {\color{blue}POOL: $2 \times 2$ size, 1 stride}
\item {\color{orange}FC: 1024 Hidden Neurons}
\item {\color{purple}DROPOUT: $p = 0.5$}
\item {\color{orange}FC: 10 Output Classes}
\end{enumerate}

At the $10^{th}$ layer of the CNN, instead of the conventional softmax function with the cross entropy function (for computing loss), the L2-SVM is implemented. That is, the output shall be translated to the following case $y \in \{-1, +1\}$, and the loss is computed by Eq. \ref{l2-svm}. The \texttt{weight} parameters are then learned using \texttt{Adam}\cite{Kingma}.

\subsection{Data Analysis}
There were two parts in the experiments for this study: (1) training phase, and (2) test case. The CNN-SVM and CNN-Softmax models were used on both MNIST and Fashion-MNIST.

Only the training accuracy, training loss, and test accuracy were considered in this study.

\section{Experiments}
The code implementation may be found at https://github.com/AFAgarap/cnn-svm. All experiments in this study were conducted on a laptop computer with Intel Core(TM) i5-6300HQ CPU @ 2.30GHz x 4, 16GB of DDR3 RAM, and NVIDIA GeForce GTX 960M 4GB DDR5 GPU.
\begin{table}[!htb]
\centering
\caption{Hyper-parameters used for CNN-Softmax and CNN-SVM models.}
		\begin{tabular}{ccc}
		\toprule
		Hyper-parameters & CNN-Softmax & CNN-SVM \\
		\midrule
		Batch Size & 128 & 128 \\
		Dropout Rate & 0.5 & 0.5 \\
		Learning Rate & 1e-3 & 1e-3 \\
		Steps & 10000 & 10000 \\
		SVM C & N/A & 1 \\
		\bottomrule
		\end{tabular}\\
		\label{table: hyperparameters}
\end{table}

The hyper-parameters listed in Table \ref{table: hyperparameters} were manually assigned, and were used for the experiments in both MNIST and Fashion-MNIST.

\begin{figure}[!htb]
\minipage{0.5\textwidth}
\centering
	\includegraphics[width=\linewidth]{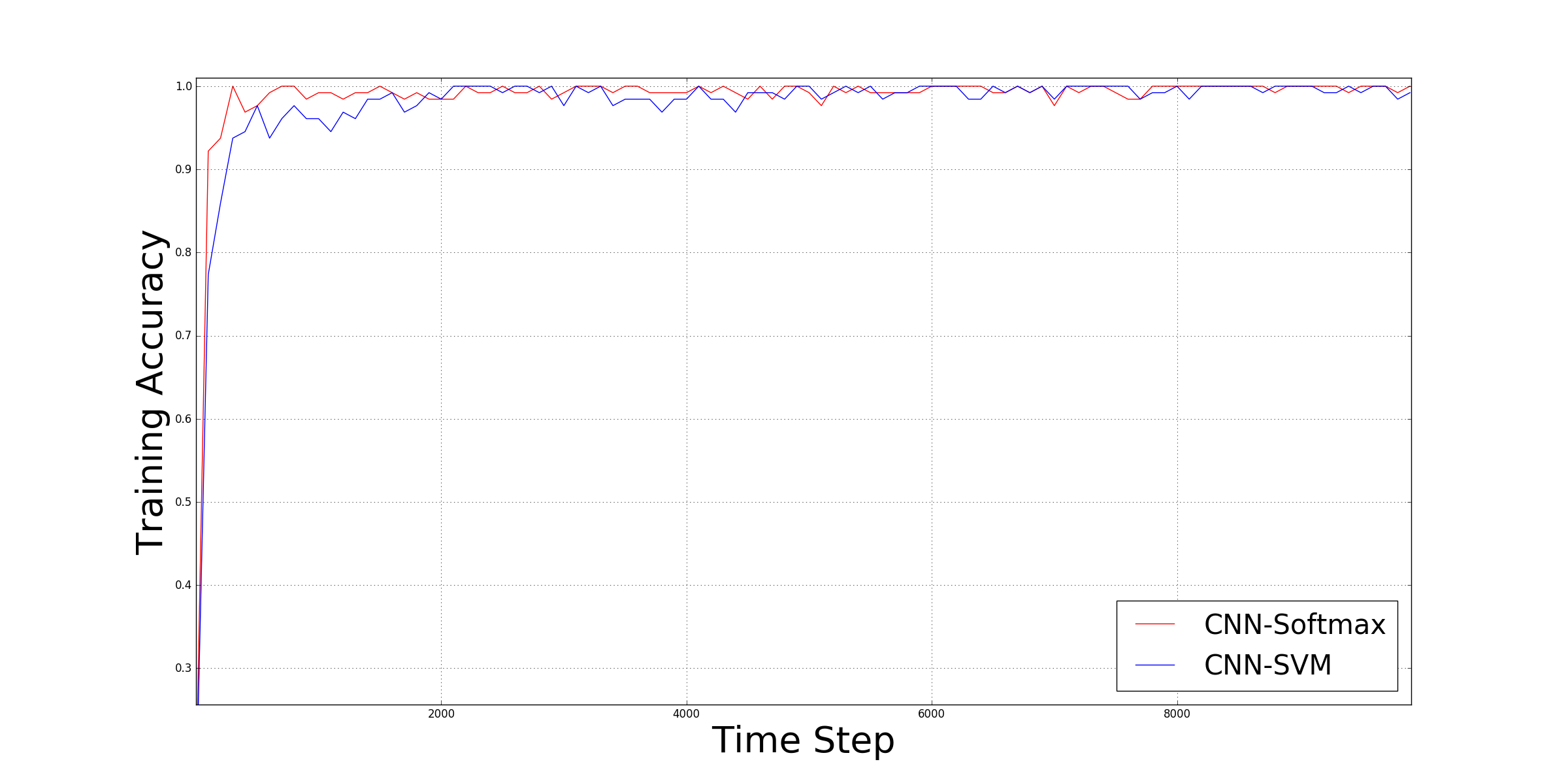}
	\caption{Plotted using \texttt{matplotlib}\cite{Hunter:2007}. Training accuracy of CNN-Softmax and CNN-SVM on image classification using MNIST\cite{lecun2010mnist}.}
	\label{training-accuracy}
\endminipage\hfill
\end{figure}

\begin{figure}[!htb]
\minipage{0.5\textwidth}
\centering
	\includegraphics[width=\linewidth]{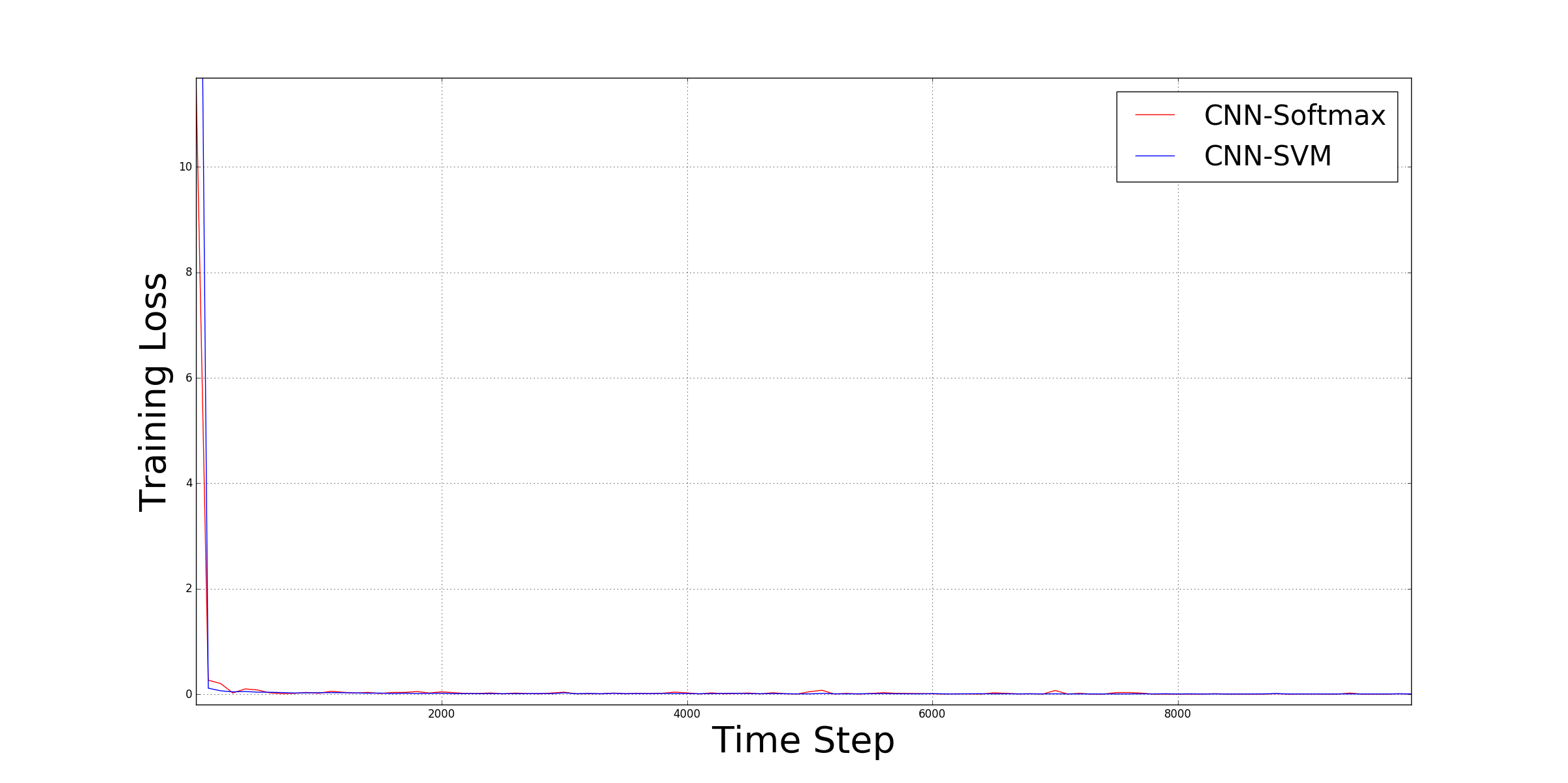}
	\caption{Plotted using \texttt{matplotlib}\cite{Hunter:2007}. Training loss of CNN-Softmax and CNN-SVM on image classification using MNIST\cite{lecun2010mnist}.}
	\label{training-loss}
\endminipage\hfill
\end{figure}

Figure \ref{training-accuracy} shows the training accuracy of CNN-Softmax and CNN-SVM on image classification using MNIST, while Figure \ref{training-loss} shows their training loss. At 10,000 steps, both models were able to finish training in 4 minutes and 16 seconds. The CNN-Softmax model had an average training accuracy of 98.4765625\% and an average training loss of 0.136794931, while the CNN-SVM model had an average training accuracy of 97.671875\% and an average training loss of 0.268976859.

\begin{figure}[!htb]
\minipage{0.5\textwidth}
\centering
	\includegraphics[width=\linewidth]{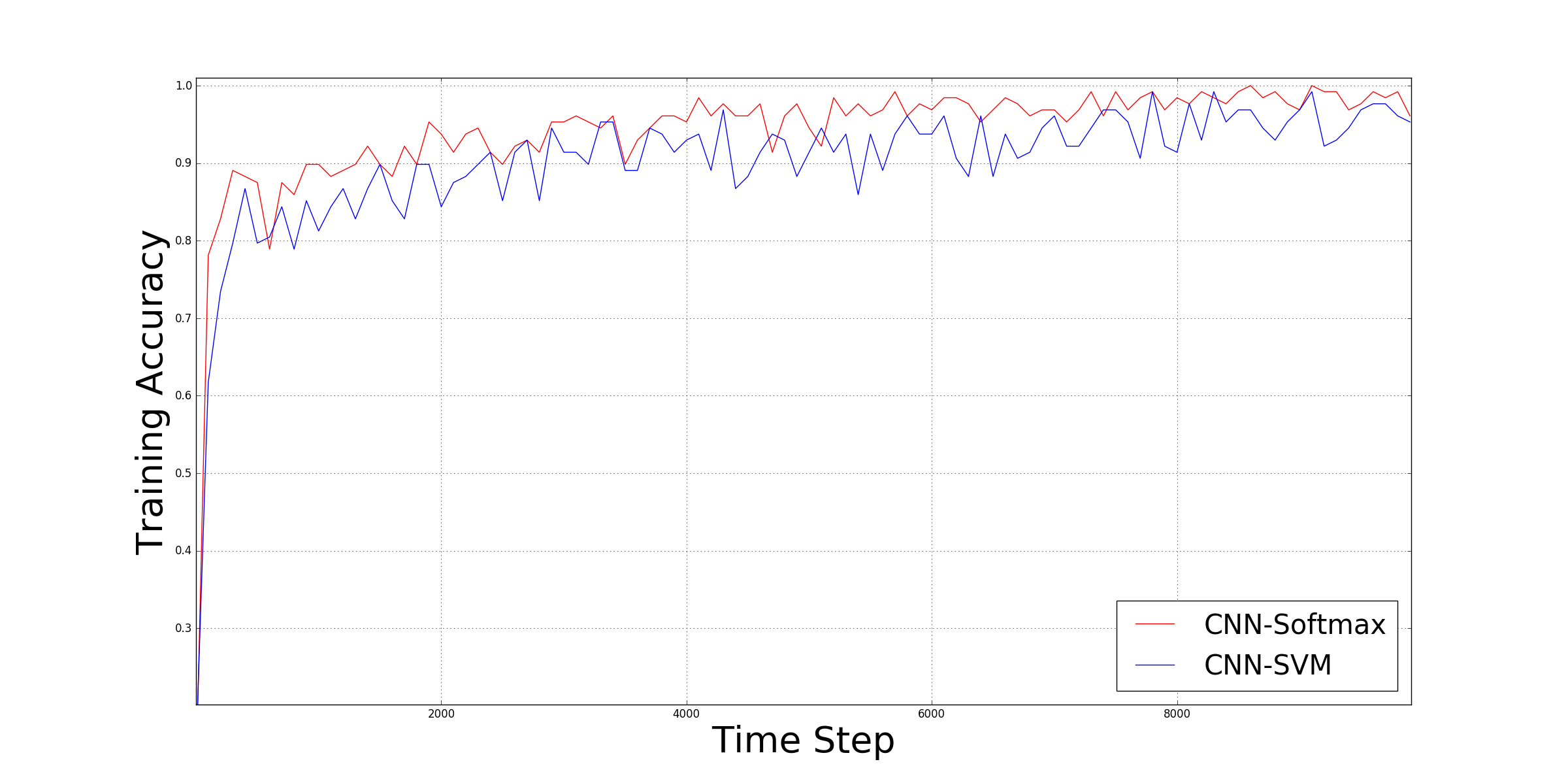}
	\caption{Plotted using \texttt{matplotlib}\cite{Hunter:2007}. Training accuracy of CNN-Softmax and CNN-SVM on image classification using Fashion-MNIST\cite{xiao2017fashion}.}
	\label{fashion-training-accuracy}
\endminipage\hfill
\end{figure}

\begin{figure}[!htb]
\minipage{0.5\textwidth}
\centering
	\includegraphics[width=\linewidth]{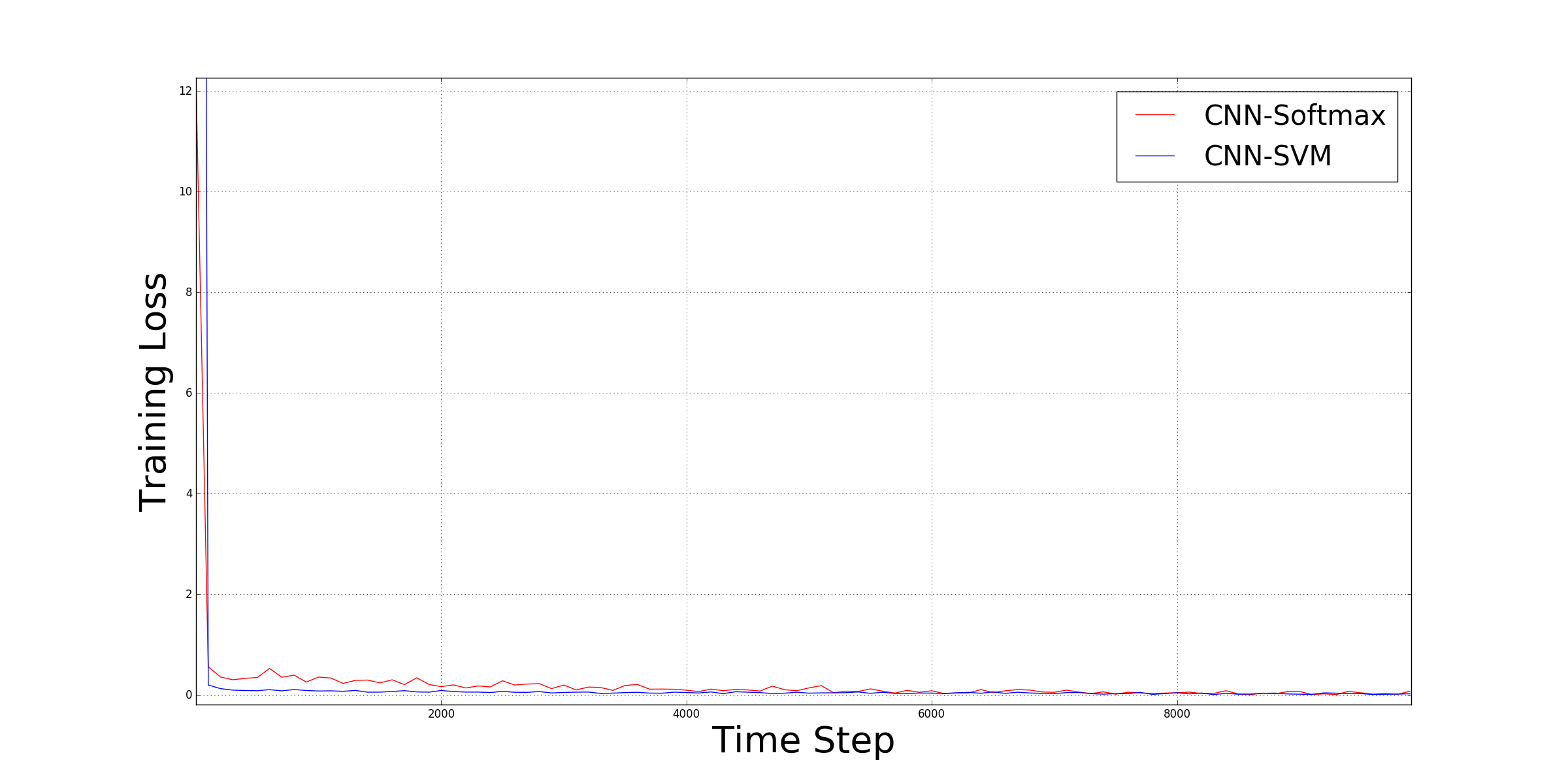}
	\caption{Plotted using \texttt{matplotlib}\cite{Hunter:2007}. Training loss of CNN-Softmax and CNN-SVM on image classification using Fashion-MNIST\cite{xiao2017fashion}.}
	\label{fashion-training-loss}
\endminipage\hfill
\end{figure}

Figure \ref{fashion-training-accuracy} shows the training accuracy of CNN-Softmax and CNN-SVM on image classification using MNIST, while Figure \ref{fashion-training-loss} shows their training loss. At 10,000 steps, the CNN-Softmax model was able to finish its training in 4 minutes and 47 seconds, while the CNN-SVM model was able to finish its training in 4 minutes and 29 seconds. The CNN-Softmax model had an average training accuracy of 94\% and an average training loss of 0.259750089, while the CNN-SVM model had an average training accuracy of 90.15\% and an average training loss of 0.793701683.

\begin{table}[!htb]
\centering
\caption{Test accuracy of CNN-Softmax and CNN-SVM on image classification using MNIST\cite{lecun2010mnist} and Fashion-MNIST\cite{xiao2017fashion}.}
		\begin{tabular}{ccc}
		\toprule
		Dataset & CNN-Softmax & CNN-SVM \\
		\midrule
		MNIST & 99.23\% & 99.04\% \\
		Fashion-MNIST & 91.86\% & 90.72\% \\
		\bottomrule
		\end{tabular}\\
		\label{table: test-accuracy}
\end{table}

After 10,000 training steps, both models were tested on the test cases of each dataset. As shown in Table \ref{table: distribution}, both datasets have 10,000 test cases each. Table \ref{table: test-accuracy} shows the test accuracies of CNN-Softmax and CNN-SVM on image classification using MNIST\cite{lecun2010mnist} and Fashion-MNIST\cite{xiao2017fashion}.

The test accuracy on the MNIST dataset does not corroborate the findings in \cite{Tang}, as it was CNN-Softmax which had a better classification accuracy than CNN-SVM. This result may be attributed to the fact the there were no data pre-processing than on the MNIST dataset. Furthermore, \cite{Tang} had a relatively more sophisticated model and methodology than the simple procedure done in this study. On the other hand, the test accuracy of the CNN-Softmax model matches the findings in \cite{zalandoresearch_2017}, as both methodology did not involve data preprocessing of the Fashion-MNIST.

\section{Conclusion and Recommendation}
The results of this study warrants an improvement on its methodology to further validate its review on the proposed CNN-SVM of \cite{Tang}. Despite its contradiction to the findings in \cite{Tang}, quantitatively speaking, the test accuracies of CNN-Softmax and CNN-SVM are almost the same with the related study. It is hypothesized that with data preprocessing and a relatively more sophisticated base CNN model, the results in \cite{Tang} shall be reproduced.

\section{Acknowledgment}
An expression of gratitude to Yann LeCun, Corinna Cortes, and Christopher J.C. Burges for the MNIST dataset\cite{lecun2010mnist}, and to Han Xiao, Kashif Rasul, and Roland Vollgraf for the Fashion-MNIST dataset\cite{xiao2017fashion}.

\bibliographystyle{ACM-Reference-Format}
\bibliography{paper} 

\end{document}